\documentclass{article}
\usepackage{subcaption}
\usepackage{booktabs}
\usepackage{multirow}
\usepackage{graphicx}
\usepackage{wrapfig}
\usepackage{caption}
\usepackage{xcolor}
\usepackage{amsmath}
\usepackage{amssymb}
\usepackage{bbding}
\usepackage{xspace}
\usepackage{xcolor}
\usepackage{float}
\usepackage{placeins}
\definecolor{softgreen}{RGB}{60,140,90}
\definecolor{softred}{RGB}{180,90,90}
\usepackage[final]{corl_2026} 
\newcommand{\alias}{ELAN4D\xspace}
\title{\alias: Embodiment-Centric 4D Supervision for Vision-Language-Action Models via Plug-and-Play Adaptation}

\author{\mdseries
Zeyuan He$^{*1,2}$, Bowen Yang$^{*4}$, Zhirui Fang$^{*\dagger3}$, Keru Zhou$^{*3}$, \\
Lei Jiang$^{\ddagger5}$, Jingjing Qian$^{2}$, Fan Mo$^{6}$, Junchi Yan$^{4}$, \\
Philip Torr$^{1}$, Xiu Li$^{3}$, Li Jiang$^{\ddagger2,8}$, Jialin Yu$^{1,7,\text{\Envelope}}$
}

\begin{document}
\maketitle
\vspace{-3em}
\begin{center}
$^1$Torr Vision Group, University of Oxford\\
$^2$The Chinese University of Hong Kong, Shenzhen\\
$^3$Tsinghua University\\
$^4$Shanghai Jiao Tong University\\
$^5$University College London\\
$^6$University of Cambridge\\
$^7$University of Bath\\
$^8$Shenzhen Loop Area Institute\\
$^*$Equal Contribution \qquad
$^\dagger$Project Lead \qquad
$^\ddagger$Equal Supervision \qquad
\textsuperscript{\Envelope}Corresponding Author(s)
\end{center}
\vspace{1em}

\begin{abstract}
    Vision-Language-Action (VLA) models have shown promise for robotic manipulation, yet most existing policies operate reactively by directly regressing actions from current observations, without explicitly modeling future dynamics. This limits their ability to generalize under out-of-distribution perturbations. To address this issue, we propose \alias, an embodiment-centric, 4D-aware training framework that enhances VLA policies with future robot keypoint tracks as predictive spatio-temporal supervision. Using only forward kinematics from proprioceptive states, we derive 3D displacement tracks of robot keypoints, such as joints and the end-effector, with negligible preprocess cost. These tracks provide metric and compact supervision without requiring external trackers or reconstruction. A plug-and-play auxiliary branch with a lightweight track decoder injects this 4D signal into the action expert, without directly propagating the auxiliary 4D objective into the pretrained vision-language pathway. The track decoder is discarded during inference, leaving the base policy interface unchanged. Extensive experiments on LIBERO, LIBERO-Plus, RoboTwin2.0 and real-world manipulation tasks demonstrate that \alias consistently improves over strong VLA baselines, achieving the best overall performance and substantial gains under out-of-distribution perturbations, including camera, background, and layout shifts. These results highlight the effectiveness of embodiment-centric 4D supervision for building more robust and generalizable manipulation policies.
\end{abstract}

\keywords{Robotic manipulation, Imitation learning, Vision-Language-Action models, 4D prediction} 

\begin{figure}[htbp]
    \centering
    \includegraphics[width=\linewidth]{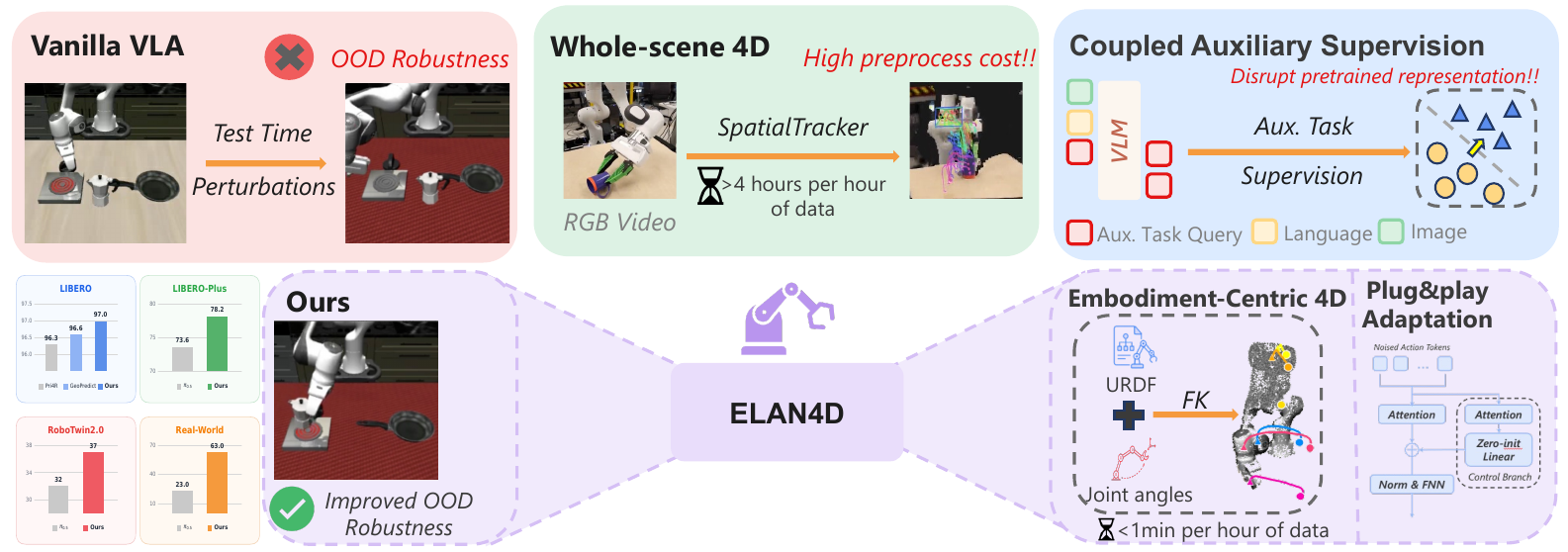}
    \caption{We present \textbf{ELAN4D}, a training framework that improves VLA policies with embodiment-centric 4D supervision via plug-and-play adaptation. ELAN4D consistently improves success rates across simulation and real-world tasks especially in out-of-domain settings.}
    \label{fig:teaser}
    \vspace{-5mm}
\end{figure}

\section{Introduction}

Recent advances in Vision-Language-Action (VLA) models have established them as a promising framework for robotic manipulation, where pretrained Vision-Language Models (VLMs) are adapted to predict robot actions from visual observations and language instructions~\citep{brohan2022rt, zitkovich2023rt, black2410pi0, kim2025openvla, kim2025fine}. These models benefit from rich semantic and visual priors learned from internet-scale data, enabling promising performance across diverse manipulation tasks and environments~\citep{zhong2025survey}. Yet manipulation is fundamentally a dynamic process: success depends not only on recognizing \textit{what to do}, but also on anticipating \textit{what will happen} while doing it. However current policies typically predict action from the current observation reactively without explicitly modeling the future dynamics induced by these actions, limiting their robustness under out-of-distribution visual and spatial shifts~\citep{li2025surveyvisionlanguageactionmodelsembodied}.

A natural response to this limitation is to supervise the policy with \textbf{predictive objectives} that force it to anticipate the future, not just imitate the next action. Early efforts pursue this goal in the 2D image space, training the policy to forecast future RGB frames~\citep{cen2025worldvla, wu2024unleashing} or depth~\citep{zhangdreamvla} alongside action prediction. While such 2D signals are easy to obtain, they remain tied to appearance-level cues, with much of the supervision coming from static background context or nuisance visual variation rather than action-relevant changes~\citep{fei2025liberoplus, chen2025robotwin}. Recent works such as GeoPredict~\citep{qian2025geopredict} and Pri4R~\citep{kim2026pri4rlearningworlddynamics} address this limitation by supervising the policy with future 3D point tracks (4D), but at substantial cost: they either depend on dense tracks from external spatial trackers~\citep{xiao2025spatialtracker}, inflating preprocessing overhead~\citep{kim2026pri4rlearningworlddynamics}, or task the VLM with predicting tracks through extra queries~\citep{qian2025geopredict}, disrupting pretrained vision-language representations which may harm generalization~\citep{zhang2026vlmvla, driess2025knowledgeinsulatingvisionlanguageactionmodels}.

These limitations suggest three design principles for practical 4D supervision in VLA policies.  First, the 4D signal should be compact and easy to obtain. In many tabletop manipulation settings, much of the scene is static, while the most reliable and densely available motion signal comes from the robot embodiment itself. \emph{Future robot keypoint tracks}, the 3D trajectories of points anchored to the robot body, therefore provide a compact embodiment-centric 4D signal that can be computed from proprioceptive states via forward kinematics without external trackers. Second, this supervision should be injected without disrupting the pretrained VLM backbone or base policy, motivating lightweight auxiliary pathways with gradient isolation. Third, the 4D signal should be introduced at training-time only, leaving the policy's input-output interface unchanged at inference.

In this paper, we propose \textbf{ELAN4D}, a training framework that improves VLA policies with embodiment-centric 4D supervision via plug-and-play adaptation. As in Figure~\ref{fig:teaser}, we represent the robot body with a sparse set of 3D keypoints placed on the arm joints and end-effector, and derive their future displacement tracks from proprioceptive trajectories. These tracks provide metric, temporally dense 4D supervision without requiring external point tracking or dense scene reconstruction. To incorporate this supervision without disrupting the pretrained VLM or base policy, ELAN4D uses a plug-and-play ControlNet-style~\citep{zhang2023controlnet} auxiliary branch with a lightweight track decoder for predicting future robot keypoint displacements. This design avoids directly perturbing the backbone with the auxiliary objective while allowing the learned 4D-aware features to support action prediction through a residual pathway. At inference time, the track decoder is discarded, preserving the base policy's input-output interface and requiring no additional queries or future predictions. Together, these choices treat embodiment-centric future prediction as a \textit{lightweight inductive bias} for action learning, rather than requiring the policy to model full scene dynamics.

Through extensive experiments on LIBERO~\citep{liu2023libero}, LIBERO-Plus~\citep{fei2025liberoplus}, RoboTwin2.0~\citep{chen2025robotwin} and real-world manipulation tasks, we show that ELAN4D consistently improves base VLA across single-arm and bimanual tasks and performs favorably against SOTA VLA methods. Its improvements are particularly strong in out-of-distribution settings, including viewpoint, background, and layout shifts, demonstrating the effectiveness of embodiment-centric 4D supervision for policy learning.

In summary, our contributions are threefold. \textbf{First}, we introduce \textbf{ELAN4D}, a VLA framework for learning 4D-aware policies from future robot keypoint tracks. \textbf{Second}, we use these tracks as compact embodiment-centric 4D supervision and inject them through a ControlNet-style action branch with gradient isolation, while preserving the policy's inference interface. \textbf{Third}, we demonstrate that ELAN4D delivers consistent improvements across simulation and real-world manipulation tasks, particularly in out-of-distribution settings requiring spatial generalization and visual robustness.


\section{Related Work}
\subsection{Vision-Language-Action Models}

Vision--Language--Action (VLA) models extend pretrained vision--language models to robotic control by predicting actions conditioned on visual observations and language instructions~\citep{kim2025openvla, black2410pi0, black2025pi, bu2025univla, zhangdreamvla, cen2025worldvla, li2025controlvla, jia2026guidedvla}. OpenVLA-style methods~\citep{kim2025openvla, bu2025univla} formulate action generation as autoregressive prediction over discrete action tokens, whereas the $\pi$ series~\citep{black2410pi0, black2025pi} adopts continuous regression and flow matching for more precise and efficient action generation. Despite this progress, many VLA policies remain largely 2D-centric~\citep{zitkovich2023rt, huvideo, brohan2022rt}, limiting their ability to reason about the 3D structure required for complex manipulation. Recent methods~\citep{li2025bridgevla, fan2026any3d, qu2025spatialvla, sun2025geovla} incorporate explicit 3D information, such as depth or point clouds, to improve spatial understanding. However, these approaches often require additional 3D inputs at inference time and primarily focus on static geometry rather than 3D dynamics. ELAN4D addresses this limitation by introducing 4D-aware auxiliary supervision during training while keeping the policy interface unchanged at inference time.

\subsection{Predictive Supervision for Robotic Manipulation}

Recent studies improve robot policies through future prediction~\citep{hafnerdream, cen2025worldvla, xu2026futurevlajointvisuomotorprediction, jiang2025rynnvla001usinghumandemonstrations, fan2026futurevlaforecastingunifiedtrajectories, hafner2022deep, qian2026escapeepisodicspatialmemory, chen2026oasisobservationactionspacealignment}, including forecasting future RGB observations~\citep{wu2024unleashing, cen2025worldvla} or depth~\citep{zhangdreamvla}. However, these objectives are often 2D-centric and provide limited explicit spatial structure~\citep{cen2025worldvla}. Moreover, single-frame prediction may be insufficient to capture the physical dynamics needed for precise manipulation~\citep{du2023learning, black2023zero}. Recent methods such as Pri4R~\citep{kim2026pri4rlearningworlddynamics} and GeoPredict~\citep{qian2025geopredict} address this issue by using predictive supervision from 3D point tracks, i.e., 4D trajectories. Nevertheless, Pri4R relies on global tracks extracted by spatial trackers~\citep{xiao2025spatialtracker}, leading to a time-consuming preprocessing pipeline. GeoPredict injects predictive supervision into the VLM through additional track queries, coupling pretrained vision--language representations with low-level dynamics forecasting, which may interfere with the VLM's native capabilities~\citep{zhang2026vlmvla, driess2025knowledgeinsulatingvisionlanguageactionmodels}. In contrast, ELAN4D introduces embodiment-centric 4D supervision into the action expert through a ControlNet-style pathway with gradient isolation, isolating the VLM representation from the auxiliary 4D objective while incurring negligible preprocessing cost.
\vspace{-1em}
\section{Method}\vspace{-1em}

\begin{figure}[t]
    \centering
    \includegraphics[width=\linewidth]{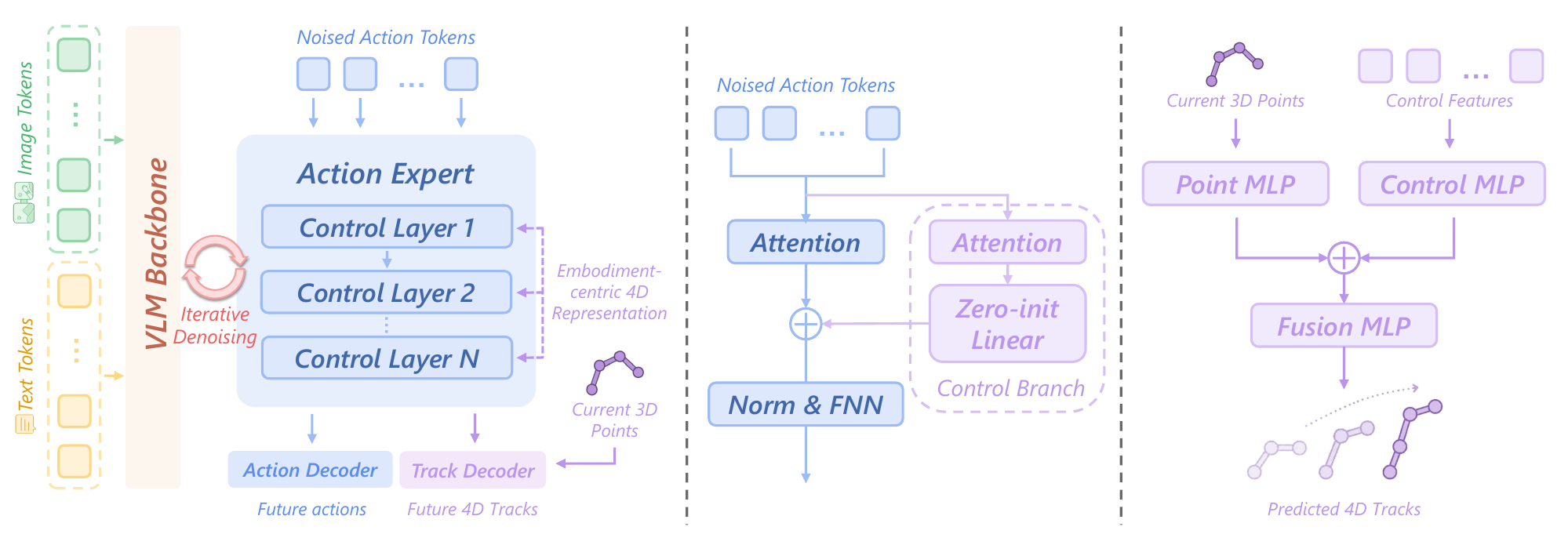}
    \caption{\textbf{Overview of ELAN4D.}
    \textbf{Left:} Image and instruction tokens are encoded by a pretrained VLM backbone and, together with noised action tokens, fed into an \emph{Action Expert} whose layers are augmented with \emph{Control Layers}. An \emph{Action Decoder} predicts future actions, while a \emph{Track Decoder} predicts future robot 3D keypoint trajectories as 4D supervision.
    \textbf{Mid:} Zoom-in of a Control Layer. A residual control branch (purple) applies attention followed by a \emph{zero-initialized linear} layer, and adds its output to the main action pathway (blue) via $\oplus$. \textbf{Right:}  Zoom-in of Track Decoder. At the final control layer, the control branch features, conditioned on the current 3D keypoint positions, feed the Track Decoder. At inference, the 3D input and Track Decoder are discarded.} 
    \label{fig:Model_Architecture}
    \vspace{-1em}
\end{figure}

We propose ELAN4D, a training framework that improves VLA policies with embodiment-centric 4D supervision. The central design is to expose the action module to 4D dynamics during training through a plug-and-play residual branch, while keeping track-loss gradients away from the pretrained vision-language pathway and preserving policy interface. ELAN4D consists of two components: a 4D supervision signal derived from proprioceptive trajectories, and a ControlNet-style branch that absorbs this auxiliary supervision through a residual pathway~\citep{zhang2023controlnet, li2025controlvla}. We first review the VLA setup (Sec.~\ref{sec:prelim}), then describe the 4D supervision signal (Sec.~\ref{sec:4d_signal}), introduce the control branch and track decoder (Sec.~\ref{sec:control_branch}), and finally present training and inference (Sec.~\ref{sec:training}).

\subsection{Preliminary}
\label{sec:prelim}
\paragraph{Problem Formulation.} We consider language-conditioned manipulation under the Vision-Language-Action (VLA) paradigm~\citep{kim2025openvla,black2410pi0}. At each step $t$, the policy takes a language instruction $\mathbf{L}$, multi-view images $\mathbf{I}_t$, and proprioceptive state $\mathbf{q}_t$ as input, and predicts an action chunk $\mathbf{A}_t = [\mathbf{a}_t, \dots, \mathbf{a}_{t+H-1}]$ of horizon $H$. Each $\mathbf{a}_t \in \mathbb{R}^7$ denoting a 7-DoF end-effector command $\mathbf{a}_t = [\Delta \mathbf{x}_t, \Delta \boldsymbol{\theta}_t, g_t]$, where $\Delta \mathbf{x}_t \in \mathbb{R}^3$ and $\Delta \boldsymbol{\theta}_t \in \mathbb{R}^3$ are the translational and rotational offsets, and $g_t \in \mathbb{R}$ is the gripper's open-close state.\vspace{-1em}
\paragraph{Base Models.} We build on the OpenPI series ($\pi_0$~\citep{black2410pi0} and $\pi_{0.5}$~\citep{black2025pi}), which combine a PaliGemma~\citep{beyer2024paligemma} VLM backbone with an action expert that predicts continuous action chunks via conditional flow matching. Our goal is to enhance action learning with auxiliary 4D supervision during training, without altering the policy interface at inference.

\subsection{Embodiment-centric 4D Supervision}
\label{sec:4d_signal}
As illustrated in Figure~\ref{fig:Model_Architecture}, ELAN4D uses future robot keypoint tracks as an auxiliary supervision signal. Specifically, we represent the robot body with a sparse set of 3D keypoints placed on the arm joints and end-effector, and supervise their displacements over the action horizon. This provides embodiment-centric 4D supervision that is naturally aligned with action generation.\vspace{-1em}

\paragraph{Track construction.}
For each demonstration trajectory, we access the proprioceptive state \(\mathbf{q}_t\) at every control step. Let \(\mathcal{K}=\{1,\dots,K\}\) denote the selected robot keypoints, including the major arm joints and the end-effector. Using the known robot kinematic chain~\citep{craig2009introduction}, forward kinematics maps each keypoint \(k\) to its Cartesian position \(\mathbf{p}_{t}^{k}\in\mathbb{R}^{3}\) in the robot base frame: \(\mathbf{p}_{t}^{k} = \mathrm{FK}_{k}(\mathbf{q}_{t})\). Let \(\mathbf{P}_{\tau}=[\mathbf{p}_{\tau}^{1}, \dots, \mathbf{p}_{\tau}^{K}] \in \mathbb{R}^{K\times 3}\) denote the full keypoint set at time \(\tau\). This signal is occlusion-free and much cheaper to obtain than video-based trackers such as SpatialTracker~\citep{xiao2025spatialtracker} (\(\sim\!1\) CPU-minute vs.\ \(>\!4\) GPU-hours per hour of data).\vspace{-1em}

\paragraph{Future displacement target.}
At time \(t\), we define the 4D supervision target as the future displacement track of robot keypoints over the action horizon. Specifically, each future keypoint set is expressed relative to the current geometry as \(\Delta \mathbf{P}_{t+h} = \mathbf{P}_{t+h} - \mathbf{P}_{t}\), for \(h=1,\dots,H\). We collect these relative displacements into
\begin{equation}
    \mathbf{Y}_{t}
    =
    \left[
    \Delta \mathbf{P}_{t+1}, \Delta \mathbf{P}_{t+2}, \dots, \Delta \mathbf{P}_{t+H}
    \right]
    \in \mathbb{R}^{H\times K\times 3}.
    \label{eq:track-target}
\end{equation}
This target specifies the robot's 3D motion over time and serves as training-time 4D supervision. While whole-scene dynamics may contain richer signals, robot keypoint tracks offer a cheaper and focused supervision signal over manipulation-relevant motion.

\subsection{ControlNet-Style Action Branch}
\label{sec:control_branch}

We next describe how embodiment-centric 4D supervision is introduced into the policy. As coupling the 4D prediction task too tightly to the VLM may disturb pretrained visual-language representations, ELAN4D attaches a trainable residual branch to the action expert, thereby confining the auxiliary supervision to the main action generation pathway.\vspace{-1em}

\paragraph{Residual control branch.}
Let \(\mathbf{u}_{t}\) denote the action-expert feature produced by the policy from the language, image, and proprioceptive inputs. ELAN4D adds a ControlNet-style branch and fuses it with the main action feature through a projection initialized to zero:
\begin{equation}
\widetilde{\mathbf{u}}_{t} = \mathbf{u}_{t} + \mathrm{Proj}(\mathbf{C}_{t}), \qquad \mathbf{C}_{t} = b_{\psi}(\mathrm{sg}(\mathbf{u}_{t})).
\end{equation}
Here \(b_{\psi}\) is the trainable control branch, \(\mathbf{C}_{t}\) denotes its output token features, and \(\mathrm{sg}(\cdot)\) denotes stop-gradient. The stop-gradient operation
prevents the auxiliary 4D objective from propagating into the pretrained vision-language representation. The projection \(\mathrm{Proj}(\cdot)\) is initialized to zero before fusion,
so the control branch initially contributes no residual signal, preserving the pretrained action behavior and stabilizing early post-training.\vspace{-1em}

\paragraph{Track decoder.}
To anchor the auxiliary prediction to the current robot state, we augment the control branch with a lightweight point-conditioned track decoder that predicts future 4D displacements for robot keypoints. At time \(t\), a point MLP embeds the current robot keypoints \(\mathbf{P}_t \in \mathbb{R}^{K \times 3}\) into per-keypoint features, while a control MLP maps the control-branch features \(\mathbf{C}_t \in \mathbb{R}^{H \times d}\) into per-step control features. We broadcast the control features across keypoints and the keypoint features across the horizon, concatenate them to form a tensor in \(\mathbb{R}^{H \times K \times (d + d_p)}\), and feed the result to a fusion MLP with residual blocks to predict per-step 3D displacements:
\begin{equation}
\widehat{\mathbf{Y}}_t = \mathrm{MLP}_{\mathrm{fusion}} \!\left( \mathrm{MLP}_{\mathrm{ctrl}}(\mathbf{C}_t) \oplus \mathrm{MLP}_{\mathrm{point}}(\mathbf{P}_t) \right) \in \mathbb{R}^{H \times K \times 3}.
\end{equation}
By conditioning displacement prediction on both current keypoints and horizon-wise control features, track decoder supervises \(\mathbf{C}_t\) to capture future robot keypoint motion. 

\subsection{Training and Inference}
\label{sec:training}
We train ELAN4D by jointly optimizing the original action objective \(\mathcal{L}_{\mathrm{act}}\), instantiated as conditional flow matching for the \(\pi\)-series base models, and an
auxiliary 4D prediction objective.\vspace{-1em}

\paragraph{Track prediction loss.}
The track loss supervises predicted robot keypoint displacements over all future steps and selected keypoints. Let \(\widehat{\Delta\mathbf{p}}_{t+h}^{k}\) and \(\Delta\mathbf{p}_{t+h}^{k}\) denote the \(k\)-th predicted and target displacements in \(\widehat{\mathbf{Y}}_t\) and \(\mathbf{Y}_t\), respectively:
\begin{equation}
    \mathcal{L}_{\mathrm{track}} = \frac{1}{HK} \sum_{h=1}^{H}\sum_{k=1}^{K} \left\| \widehat{\Delta\mathbf{p}}_{t+h}^{k} - \Delta\mathbf{p}_{t+h}^{k} \right\|_{1}.
\end{equation}
We use an \(\ell_1\) distance by default for robustness to occasional noisy state estimates. By supervising every selected keypoint at every future step, this loss encourages the branch to learn predictive and dynamically consistent 4D features.\vspace{-1em}

\paragraph{Training objective.}
We optimize the combined objective \(\mathcal{L} = \mathcal{L}_{\mathrm{act}} + \lambda_{\mathrm{track}} \mathcal{L}_{\mathrm{track}}\), where \(\lambda_{\mathrm{track}}\) balances imitation learning and 4D prediction. The two losses are applied to different parameter subsets. \(\mathcal{L}_{\mathrm{act}}\) supervises the main action-generation pathway and the added control branch, while \(\mathcal{L}_{\mathrm{track}}\) updates only the control branch and the track decoder. A stop-gradient operation at the branch input prevents \(\mathcal{L}_{\mathrm{track}}\) from propagating into the pretrained VLM backbone and the original action branch. Thus, the track loss serves as an auxiliary training signal without directly modifying the pretrained visual-language representation.\vspace{-1em}

\paragraph{Inference.}
At inference time, ELAN4D discards the track decoder and does not require future robot keypoint tracks as inputs or outputs. The policy consumes the same language, image, and proprioceptive inputs as the base VLA and predicts the same action chunk, with only the learned residual control branch retained in the action expert.\vspace{-1em}
\section{Experiments}
\subsection{Experimental Setup}
\paragraph{Simulation Benchmarks.}
\textbf{LIBERO}~\citep{liu2023libero} is a Franka Emika Panda benchmark for
lifelong robotic manipulation, covering four suites: Spatial, Object, Goal, and
Long.
\textbf{LIBERO-Plus}~\citep{fei2025liberoplus} is a large-scale  benchmark
that extends LIBERO with systematically perturbed manipulation tasks. It
evaluates VLA policies under out-of-distribution simulation settings across
seven perturbation dimensions, providing a stress test for robustness and
generalization.
\textbf{RoboTwin2.0}~\citep{chen2025robotwin} is a benchmark for bimanual
manipulation that supports multi-task evaluation across diverse robot
embodiments, scenes, and objects. We use the AgileX Piper dual-arm setup and
test the policy on eight representative unseen settings to evaluate out-of-domain generalization, each task is evaluated for 100 trials. \textbf{Together}, these three benchmarks provide comprehensive evaluation of our method. LIBERO
enables comparison with recent 4D-supervised VLA methods~\citep{qian2025geopredict, kim2026pri4rlearningworlddynamics} that report results on
it but do not release code or models. LIBERO-Plus evaluates
out-of-distribution robustness in single-arm manipulation, while RoboTwin2.0
further extends the evaluation to bimanual out-of-distribution generalization. Further details about simulation benchmarks are provided in the appendix.\vspace{-1em}

\paragraph{Real-world Evaluation Suite.}
To evaluate ELAN4D on real-world tasks requiring 
spatio-temporal understanding, we design three task categories as illustrated in Figure~\ref{fig:realworld_tasks}, 
each trained with 50 expert trajectories and evaluated over 20 
trials.
\textbf{Visual Robustness:} The robot picks a fruit (alternating 
between an \emph{apple} and an \emph{orange}) and places it into a basket. At test time, the scene is populated with task-irrelevant distractors \emph{unseen} during training, testing robustness to visual distractors.
\textbf{Spatial Generalization:} The robot stacks paper cups by 
insertion. We evaluate the policy with the cups located at positions unseen during training, testing its ability to generalize to novel target locations.
\textbf{Temporal Reasoning:} The robot performs a two-stage 
assembly: Placing a cylindrical block on a base, then stacking a 
cap on top. Errors in the first stage propagate to the second, 
testing the policy's ability to chain precise manipulation without compounding errors.\vspace{-1em}

\paragraph{Baselines.} Our primary baselines are our base VLA backbone, $\pi_0$ and $\pi_{0.5}$, trained without our proposed modules. This comparison directly isolates the contribution of our 4D-aware supervision. We also compare against a range of state-of-the-art VLA approaches, including DreamVLA~\citep{zhangdreamvla}, GuidedVLA~\citep{jia2026guidedvla}, Spatial Forcing~\citep{li2026spatial}, GeoPredict~\citep{qian2025geopredict} and Pri4R~\citep{kim2026pri4rlearningworlddynamics}.\vspace{-1em}

\paragraph{Implementation Details.}
During fine-tuning, we use the original LIBERO dataset for both LIBERO and
LIBERO-Plus, which contains approximately 2K expert demonstrations collected in
simulation, without using the augmented data from LIBERO-Plus. For RoboTwin2.0,
we collect 100 expert episodes per task under the clean setting. For 4D
supervision, we track \(K=8\) keypoints for LIBERO and LIBERO-Plus (7 joints, 1 end-effector), \(K=14\) keypoints for RoboTwin2.0 (6+6 joints, 1+1 end-effector) and \(K=7\) for real world tasks. We train all models for 30K steps using AdamW (LR 2.5e-5) on 8 NVIDIA GH200 GPUs, with a total batch size of 64 and
\(\lambda_{\mathrm{track}} = 0.1\).

\subsection{Main Results}


\paragraph{LIBERO.}
Table~\ref{tab:libero_results} reports results on the original LIBERO suites. 
Despite near-saturated performance of recent VLAs on this benchmark, 
ELAN4D improves both base policies and achieves competitive overall 
success rate. ELAN4D($\pi_0$) lifts the overall success rate from 
94.2\% to 95.0\%, with the largest gain on LIBERO-Long ($+6.6$), suggesting that 4D supervision is most useful when temporal consistency matters.
ELAN4D($\pi_{0.5}$) further reaches 97.0\%, surpassing both the 
$\pi_{0.5}$ baseline and recent 4D-supervised methods such as Pri4R and 
GeoPredict. These gains indicate that our control-branch design injects 
4D predictive supervision into the base VLA effectively.\vspace{-1em}

\paragraph{LIBERO-Plus.}
Table~\ref{tab:libero_plus_results} evaluates robustness under systematic 
perturbations. ELAN4D yields substantial gains over base 
policies, raising overall success from 53.6\% to 67.6\% for $\pi_0$ and from 73.6\% to 78.2\% for $\pi_{0.5}$. The largest improvements appear on perturbations that alter visual or physical scene configurations. ELAN4D($\pi_{0.5}$) gains $+5.2$ under robot init-state and $+9.0$ under background perturbations, leading to the best overall score among all compared methods. This is consistent with ELAN4D encouraging action representations that rely less on nuisance visual and configuration cues.

\begin{table}[H]
\vspace{-1.5em}
\centering
\caption{\textbf{LIBERO-Plus benchmark results.} ELAN4D achieves the best overall performance, consistently improving over its corresponding base models, $\pi_0$ and $\pi_{0.5}$.}
\label{tab:libero_plus_results}
\resizebox{\columnwidth}{!}{%
\begin{tabular}{l ccccccc ccccc}
\toprule
\multirow{2}{*}{Method} 
& \multicolumn{7}{c}{Perturbation Dimensions} 
& \multicolumn{5}{c}{Task Suites} \\
\cmidrule(lr){2-8} \cmidrule(lr){9-13}
& Camera & Robot & Language & Light & Background & Noise & Layout 
& Spatial & Object & Goal & Long & Overall \\
\midrule
OpenVLA~\citep{kim2025openvla}              & 0.8  & 3.5  & 23.0 & 8.1  & 34.8 & 15.2 & 28.5 & 19.4 & 14.0 & 15.1 & 14.3 & 15.6 \\
OpenVLA-OFT~\cite{kim2025fine}          & 56.4 & 31.9 & \textbf{79.5} & 88.7 & \textbf{93.3} & 75.8 & 74.2 & 84.0 & 66.5 & 63.0 & 66.4 & 69.6 \\
UniVLA~\citep{bu2025univla}               & 1.8  & 46.2 & 69.6 & 69.0 & 81.0 & 21.2 & 31.9 & 55.5 & 36.7 & 40.7 & 39.9 & 42.9 \\
WorldVLA~\citep{cen2025worldvla}             & 0.1  & 27.9 & 41.6 & 43.7 & 17.1 & 10.9 & 38.0 & 32.5 & 28.6 & 31.8 & 8.2  & 25.0 \\
RIPT-VLA~\citep{tan2025RIPT-VLA}             & 55.2 & 31.2 & 77.6 & 88.4 & \underline{91.6} & 73.5 & 74.2 & \underline{85.8} & 64.3 & 58.0 & \underline{67.5} & 68.4 \\
DreamVLA~\citep{zhangdreamvla}             & \underline{65.0} & 40.8 & 63.5 & 85.7 & 82.6 & \underline{84.9} & 74.0 & 79.7 & 79.0 & 61.7 & 59.8 & 69.9 \\
AdaMoE~\citep{shen2025adamoe}               & 53.8 & 17.5 & 20.6 & 73.7 & 73.8 & 58.6 & 65.8 & 51.0 & 57.9 & 53.3 & 38.1 & 50.1 \\
GuidedVLA~\citep{jia2026guidedvla}           & \textbf{73.7} & 51.4 & 62.6 & \textbf{94.6} & 89.0 & \textbf{85.2} & 79.9 & 84.0 & 80.9 & \underline{70.8} & 66.2 & \underline{75.4} \\
Spatial Forcing~\citep{li2026spatial}      & 20.1 & 13.4 & 40.9 & 29.1 & 33.4 & 25.7 & 39.3 & 52.9 & 31.0 & 28.2 & 5.4  & 29.1 \\
VLA-Adapter~\citep{wang2025vlaadapter}          & 36.2 & 37.9 & 74.6 & 70.6 & 76.1 & 58.0 & 69.7 & 85.0 & 46.3 & 56.0 & 50.4 & 59.1 \\
\midrule
$\pi_0$~\citep{black2410pi0}              & 13.8 & 6.0  & 58.8 & 85.0 & 81.4 & 79.0 & 68.8 & 60.7 & 61.4 & 44.9 & 48.4 & 53.6 \\
$\pi_{0.5}$~\citep{black2025pi}          & 59.7 & \underline{65.5} & 75.3 & 87.0 & 82.4 & 72.1 & \underline{80.3} & 79.9 & \textbf{87.8} & 69.0 & 64.9 & 73.6 \\
ELAN4D($\pi_0$)     & 61.8 & 38.4 & 60.6 & 89.1 & 84.1 & 77.8 & 72.1 & 78.8 & 74.0 & 62.6 & 55.9 & 67.6 \\
ELAN4D($\pi_{0.5}$) & 63.7 & \textbf{70.7} & \underline{77.8} & \underline{89.8} & 91.4 & 79.9 & \textbf{81.4} & \textbf{86.8} & \underline{84.5} & \textbf{71.5} & \textbf{70.3} & \textbf{78.2} \\
\bottomrule
\end{tabular}%
}
\vspace{-2mm}
\end{table}

\begin{minipage}[t]{0.58\columnwidth}
    \vspace{0pt}
    \centering

    \captionof{table}{\textbf{LIBERO benchmark results.} ELAN4D achieves the best overall performance, competitive with SOTA models using 4D supervision.}
    \label{tab:libero_results}


    \begin{minipage}[t][3.2cm][t]{\linewidth}
        \centering
        \scriptsize
        \setlength{\tabcolsep}{2.5pt}
        \resizebox{\linewidth}{!}{%
        \begin{tabular}{lccccc}
        \toprule
        Model & Spatial & Object & Goal & Long & Overall \\
        \midrule
        $\pi_0$~\citep{black2410pi0}           & 96.8 & \underline{98.8} & 95.8 & 85.2 & 94.2 \\
        $\pi_{0.5}$~\citep{black2025pi}        & \textbf{98.8} & 98.2 & \underline{98.0} & 92.4 & \underline{96.9} \\
        Pri4R~\citep{kim2026pri4rlearningworlddynamics} & 93.2 & 98.6 & \textbf{98.1} & \textbf{95.3} & 96.3 \\
        GeoPredict~\citep{qian2025geopredict}  & 98.0 & 98.2 & 95.7 & 94.0 & 96.6 \\
        ELAN4D($\pi_0$)                  & 96.4 & 98.2 & 93.4 & 91.8 & 95.0 \\
        ELAN4D($\pi_{0.5}$)              & \underline{98.2} & \textbf{98.8} & 96.8 & \underline{94.2} & \textbf{97.0} \\
        \bottomrule
        \end{tabular}%
        }
        \vfill
    \end{minipage}
\end{minipage}
\hfill
\begin{minipage}[t]{0.38\columnwidth}
    \vspace{0pt}
    \centering

    \begin{minipage}[t][3.2cm][t]{\linewidth}
        \centering
        \includegraphics[width=\linewidth,height=3.2cm,keepaspectratio]{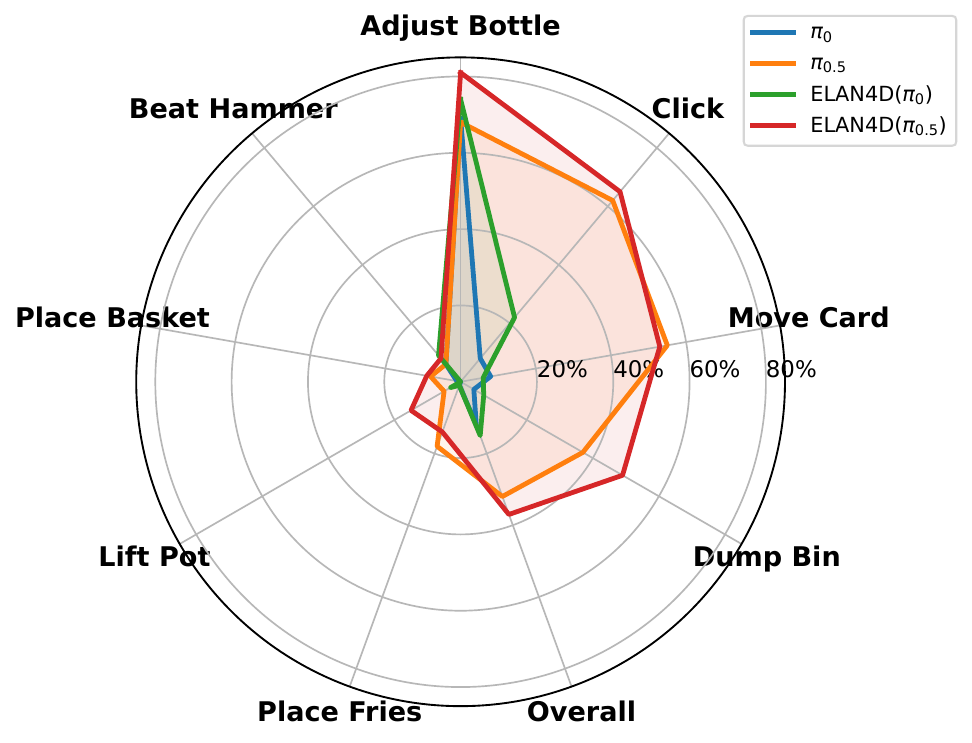}
        \vfill
    \end{minipage}

    \vspace{0.4em}

    \captionof{figure}{\textbf{RoboTwin2.0 benchmark results.} ELAN4D consistently improves over its base models.}
    \label{fig:radar_success_rate}
\end{minipage}

\paragraph{RoboTwin2.0.}Figure~\ref{fig:radar_success_rate} further evaluates ELAN4D's robustness under unseen bimanual settings.
ELAN4D improves the overall success rate of \(\pi_0\) from \(12\%\) to
\(15\%\), and improves \(\pi_{0.5}\) from \(32\%\) to \(37\%\). The gains are
especially clear on tasks requiring spatial understanding, such as
\textit{Adjust Bottle}, \textit{Dump Bin}, and \textit{Lift Pot}. For
ELAN4D(\(\pi_{0.5}\)), success on \textit{Dump Bin} increases from \(37\%\) to
\(49\%\), and \textit{Lift Pot} improves from \(5\%\) to \(15\%\). These results suggest that ELAN4D remains effective in more challenging
bimanual settings. Success rates corresponding to Figure~\ref{fig:radar_success_rate} are presented in Appendix.
\vspace{-1em}


\subsection{Real-World Experiments}

\begin{figure}[H]
    \vspace{-2em}
    \centering
    \begin{subfigure}[b]{0.68\linewidth}
        \centering
        \includegraphics[width=\linewidth]{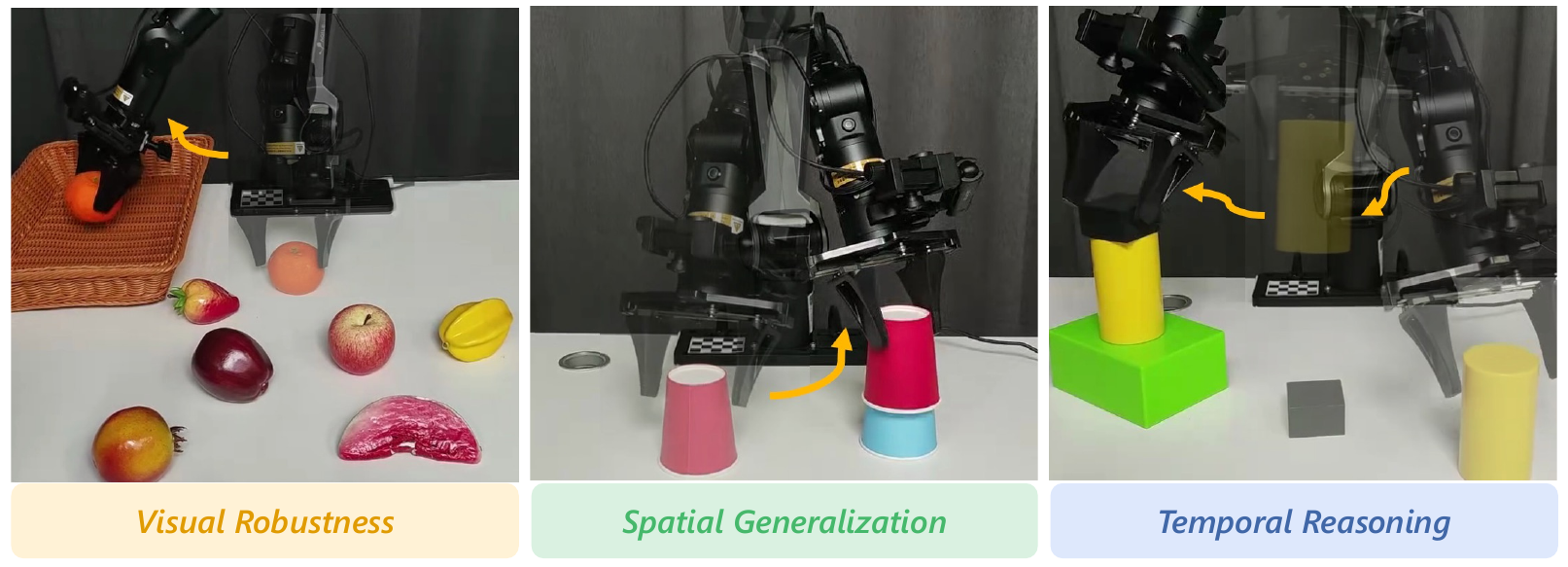}
        \caption{Illustration of Real-world task settings.}
        \label{fig:realworld_tasks}
    \end{subfigure}
    \hfill
    \begin{subfigure}[b]{0.29\linewidth}
        \centering
        \small
        \setlength{\tabcolsep}{4pt}
        \includegraphics[width=\linewidth]{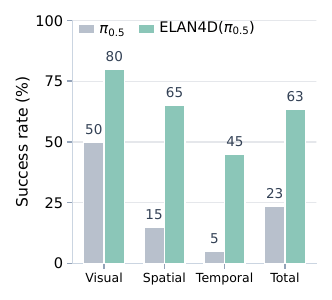}
        \caption{Task success rates (\%).}
        \label{fig:realworld_table}
    \end{subfigure}
    \vspace{-0.5em}
    \caption{\textbf{Real-world evaluation.}
    \textbf{(a)} Three real-world task settings testing visual robustness, spatial generalization, and temporal reasoning.
    \textbf{(b)} Success rates (\%) on the three real-world tasks. ELAN4D consistently improves $\pi_{0.5}$ across all three tasks.}
    \label{fig:realworld_results}
    \vspace{-2em}
\end{figure}

We evaluate ELAN4D on an AgileX Piper arm across three real-world task categories: visual robustness, spatial generalization, and temporal reasoning. As shown in Figure~\ref{fig:realworld_results}, ELAN4D consistently outperforms the $\pi_{0.5}$ baseline across all three tasks. On Robustness and Spatial tasks, ELAN4D improves success from 50\% to 80\% and from 15\% to 65\%, respectively, demonstrating stronger robustness to unseen distractors and better spatial generalization. On temporal task, ELAN4D raises success from 5\% to 45\%, suggesting that embodiment-centric 4D supervision helps reduce error accumulation in long-horizon manipulation. More illustrations are provided in the appendix.

\subsection{Ablation Study and Analysis}

\begin{figure}[t]
\centering
\begin{subfigure}[b]{0.42\textwidth}
    \centering
    \small
    \setlength{\tabcolsep}{3pt}
    \begin{tabular}{lcc}
    \toprule
    Variant & SR & $\Delta$ \\
    \midrule
    Base VLA $\pi_{0.5}$                 & 73.6 & -- \\
    + Control branch (no 4D)             & 73.3 & {\color{softred}-0.3} \\
    \midrule
    \multicolumn{3}{l}{\textit{Where to predict 4D}} \\
    \quad VLM + track queries            & 66.8 & {\color{softred}-6.8} \\
    \quad Control branch (ours)          & 78.2 & {\color{softgreen}+4.6} \\
    \midrule
    \multicolumn{3}{l}{\textit{What to predict}} \\
    \quad Whole-scene Track              & 79.3 & {\color{softgreen}+5.7} \\
    \quad Robot Keypoints (ours)         & 78.2 & {\color{softgreen}+4.6} \\
    \bottomrule
    \end{tabular}
    \caption{Ablation on LIBERO-Plus.}
    \label{tab:ablation}
\end{subfigure}
\hfill
\begin{subfigure}[b]{0.28\textwidth}
    \centering
    \includegraphics[width=\linewidth]{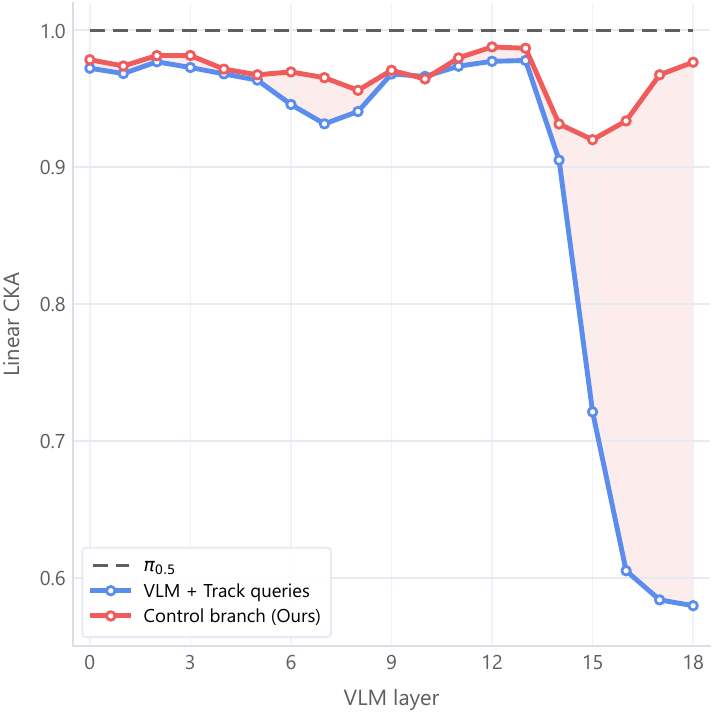}
    \caption{CKA similarity analysis.}
    \label{fig:cka}
\end{subfigure}
\hfill
\begin{subfigure}[b]{0.28\textwidth}
    \centering
    \includegraphics[width=\linewidth]{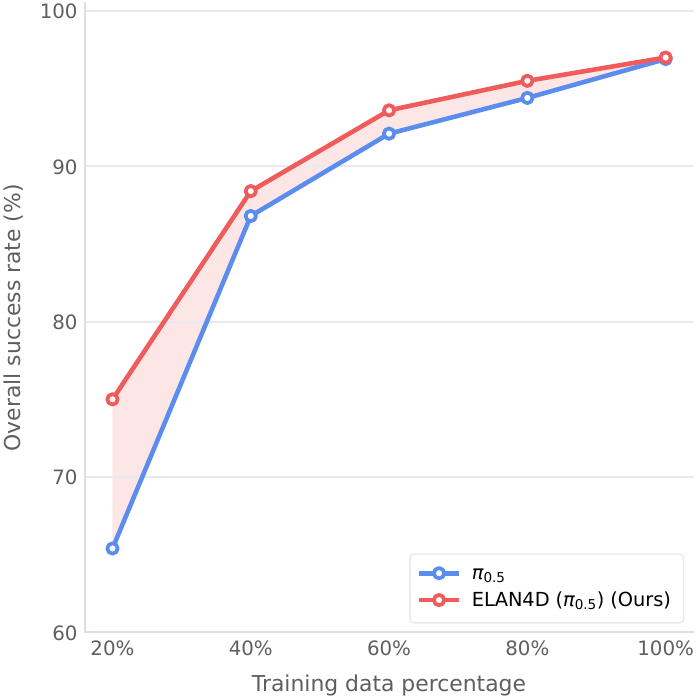}
    \caption{Data scaling on LIBERO.}
    \label{fig:data_scaling}
\end{subfigure}
\caption{\textbf{Analysis on LIBERO-Plus and LIBERO.}
\textbf{(a)} Ablation on key design choices on LIBERO-Plus. 
Gains come from 4D supervision rather than added parameters. Attaching 4D prediction to the control branch outperforms attaching it to VLM via track queries. Robot keypoint tracks perform comparably to whole-scene tracks with lower preprocessing cost.
\textbf{(b)} Layer-wise linear CKA between VLM of LIBERO-finetuned $\pi_{0.5}$ and two 4D supervised variants. Predicting 4D inside the VLM (query tokens) causes large representational drift, while our control branch keeps the VLM features close to the baseline. \textbf{(c)} Data efficiency on LIBERO: ELAN4D consistently 
outperforms the baseline $\pi_{0.5}$ across all data ratios.}
\label{fig:analysis}
\vspace{-1.5em}
\end{figure}

\paragraph{Accuracy of predicted robot tracks.}
To assess whether the auxiliary decoder learns meaningful future robot motion,
we compare its predictions with the motion realized by ELAN4D($\pi_{0.5}$)
across 10,030 LIBERO-Plus rollout episodes. Since the policy replans every
five actions, we evaluate only the executed five-step horizon of each chunk.
The decoder is enabled solely for this diagnostic evaluation; its predictions
are not fed back into action generation. ADE averages Euclidean position
error over executed steps, robot keypoints, and action chunks; FDE@5 measures
error at the fifth step, and EEF-ADE considers only the end effector.
Directional cosine similarity measures alignment between predicted and realized
stepwise motion. We also report ADE for a zero-motion predictor that holds
each keypoint at its chunk-start position.

\begin{table}[htbp]
\centering

\caption{\textbf{Robot track prediction accuracy on LIBERO-Plus.}
Predictions are compared with realized motion over the five executed steps
of each action chunk. Position errors are in centimeters; directional cosine
similarity is dimensionless. Zero-motion ADE uses the same evaluation protocol.}
\label{tab:track_accuracy}
\small
\setlength{\tabcolsep}{4pt}
\begin{tabular}{lrrrrrr}
\toprule
Rollout split & \# Episodes & ADE$\downarrow$ & FDE@5$\downarrow$
& EEF-ADE$\downarrow$ & Dir.\ cos.$\uparrow$ & Zero-motion ADE$\downarrow$ \\
\midrule
All & 10,030 & 0.734 & 1.205 & 0.799 & 0.657 & 1.095 \\
Successful & 7,823 & 0.718 & 1.186 & 0.744 & 0.737 & 1.304 \\
Failed & 2,207 & 0.757 & 1.233 & 0.881 & 0.533 & 0.782 \\
\bottomrule
\end{tabular}
\end{table}

Table~\ref{tab:track_accuracy} shows an overall ADE of 0.734\,cm,
33.0\% below the zero-motion baseline. On successful rollouts, the reduction
reaches 44.9\%, compared with 3.2\% on failed rollouts. Successful
rollouts have higher directional cosine similarity (0.737 vs.\ 0.533) and
lower EEF-ADE (0.744 vs.\ 0.881\,cm). These results associate better future-motion
alignment with successful action generation and support the interpretation
that the decoder learns action-relevant robot motion. \vspace{-1em}

\paragraph{Effect of 4D supervision.}
To isolate the contribution of 4D supervision from the added parameters of 
the control branch, we train a variant that retains the full control-branch 
architecture but removes the track prediction loss. As shown in 
Table~\ref{tab:ablation}, this variant scores 73.3\% on LIBERO-Plus, matching the base VLA $\pi_{0.5}$ (73.6\%) and far below our method (78.2\%). The gain of ELAN4D therefore comes from the embodiment-centric 4D supervision itself, not from extra capacity.\vspace{-1em}

\paragraph{Ablation on VLM-predicted 4D.}
We compare against an alternative that lets the VLM itself predict future 4D dynamics, by encoding current robot keypoints into 3D tokens and appending learnable track-query tokens to the VLM input (details are in the appendix). As shown in Table~\ref{tab:ablation}, this variant drops to 66.8\% 
($-6.8$), substantially underperforming our control-branch design (78.2\%) on tasks requiring strong generalization. We attribute this to the auxiliary query tokens and trajectory objective disrupting the VLM's pretrained representations. The Centered Kernel Alignment (CKA) analysis in Figure~\ref{fig:cka} corroborates this: relative to the 
finetuned $\pi_{0.5}$ VLM, the query-token variant shows markedly lower similarity than ours, indicating larger representational drift. Our design avoids this by isolating 4D prediction in the control branch, limiting representational drift in the VLM backbone.\vspace{-1em}

\paragraph{Are whole-scene keypoints necessary?}
We additionally consider supervising both robot and 
objects-of-interest keypoint tracks. Together with robot keypoints, these tracks capture the dominant task-relevant moving components in these scenes. To construct a stronger privileged scene-level supervision baseline, we obtain object keypoints from simulator ground-truth, avoiding any bias from off-the-shelf trackers. Even with such privileged supervision, whole-scene keypoints yield only a marginal gain over our robot-only design (1.1\%, Table~\ref{tab:ablation}). While richer scene-level signals can in principle help, this result suggests that in these scenarios, a cheap embodiment-centric 4D surrogate already captures most of the benefit without requiring expensive trackers. The cost gap is striking in practice: extracting whole-scene keypoint tracks from 1 hour of 
video via a SAM~\citep{sam} + spatial-tracker~\citep{xiao2025spatialtracker} 
pipeline takes $\sim$4 GPU-hours, whereas robot keypoint tracks can be obtained from proprioception in under 1 CPU-minute. We therefore adopt robot keypoints as our default.\vspace{-1em}

\paragraph{Performance on Data Scaling.} 
We uniformly sample 20\%, 40\%, 60\% and 80\% of the LIBERO data, train both $\pi_{0.5}$ and ELAN4D and evaluate on LIBERO (Figure~\ref{fig:data_scaling}).
ELAN4D outperforms $\pi_{0.5}$ at every budget, and the gap widens as data shrinks: with only 20\% of the data, it reaches 75.0\%, 10 percentage points above $\pi_{0.5}$, highlighting its improved data efficiency.

\section{Conclusion}
We presented ELAN4D, a training framework that improves VLA policies with embodiment-centric 4D supervision. ELAN4D uses future robot keypoint tracks as a compact 4D signal and injects this supervision through a ControlNet-style action branch with a lightweight track decoder. 
Across LIBERO, LIBERO-Plus, RoboTwin2.0, and real-world manipulation tasks, ELAN4D consistently improves strong VLA baselines, with especially clear gains under out-of-distribution perturbations. These results suggest that structured future supervision need not model the full scene to benefit VLA policies: a compact, action-aligned signal from the robot embodiment can provide an effective predictive inductive bias. \textbf{Limitations:} This design deliberately trades complete scene-level dynamics for efficient supervision, and may be less suitable for tasks dominated by external object motion, deformable objects, or complex contacts not captured by the robot embodiment. Nevertheless, our results show that this lightweight signal provides a strong and deployment-friendly training objective with negligible preprocessing cost.

\subsubsection*{Acknowledgments}

This work is supported by the Shenzhen Loop Area Institute under grant 2026NSFP006





\bibliography{references}  

\end{document}